\documentclass[letterpaper, 10 pt, conference]{ieeeconf}  

\IEEEoverridecommandlockouts                              

\usepackage{amsmath,amsfonts,bm}

\def\eqref#1{equation~\ref{#1}}

\def\1{\bm{1}}

\DeclareMathAlphabet{\mathsfit}{\encodingdefault}{\sfdefault}{m}{sl}
\SetMathAlphabet{\mathsfit}{bold}{\encodingdefault}{\sfdefault}{bx}{n}

\usepackage{graphicx}
\usepackage{subcaption}
\usepackage{amssymb}       
\usepackage{nicefrac}       
\usepackage{microtype}      
\usepackage{color}
\usepackage{booktabs}
\usepackage{url}

\title{\LARGE \bf
Learning Semantic Embedding Spaces for Slicing Vegetables
}

\author{Mohit Sharma$^{1}$, Kevin Zhang$^{1}$ and Oliver Kroemer$^{1}$
\thanks{$^{1}$Robotics Institute, 
              Carnegie Mellon University,
              Pittsburgh, PA, 15213
        {\tt\small \{mohits1, klz1, okroemer\}@cs.cmu.edu}}%
\thanks{This work was supported by Sony Corporation.}
}

\begin{document}

\maketitle
\thispagestyle{empty}
\pagestyle{empty}

\begin{abstract}
In this work, we present an interaction-based approach to learn semantically rich representations for the task of slicing vegetables.
Unlike previous approaches, we focus on object-centric representations and use auxiliary tasks to learn rich representations using a two-step process. First, we use simple auxiliary tasks, such as predicting the thickness of a cut slice, to learn an embedding space which captures object properties that are important for the task of slicing vegetables.
In the second step, we use these learned latent embeddings to learn a forward model. Learning a forward model affords us to plan online in the latent embedding space and forces our model to improve its representations while performing the slicing task. 
To show the efficacy of our approach we perform experiments on two different vegetables: cucumbers and tomatoes. 
Our experimental evaluation shows that our method is able to capture important semantic properties for the slicing task, such as the thickness of the vegetable being cut. We further show that by using our learned forward model, we can plan for the task of vegetable slicing.
\end{abstract}

\section{INTRODUCTION}

Learning suitable representations for manipulation tasks is a long-standing challenge in robotics. The complex environments of manipulation tasks usually need to be abstracted into suitable feature representations before they can be utilized by the robot to either learn skill policies or forward models. We therefore need to ensure that this abstraction process distills out the most relevant information pertaining to the task being performed. Appropriate parameterization of the feature vector may afford easier learning or even a broader generalization of the learned policies or models. Thus, acquiring a suitable representation of the environment is a crucial step towards creating an efficient learning system. 

\begin{figure}[h!]
    \centering
    \includegraphics[width=0.46\textwidth]{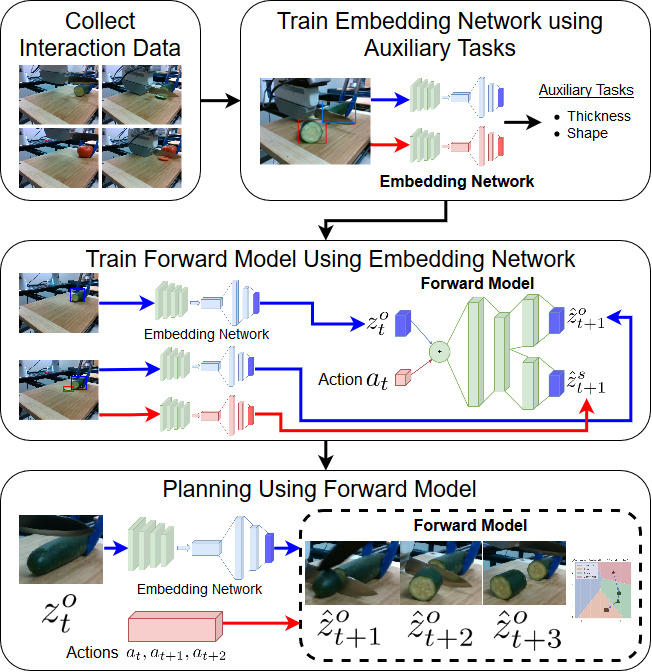}
    \caption{Overview of our framework. We first train an embedding network using auxiliary tasks such as predicting the thickness of slices. We then use the embedding network to learn a forward model that can used for multiple slicing actions}
    \label{fig:teaser_img}
    \vspace{-2em}
\end{figure}

In the past, feature vector parameterizations have often been manually predefined, even if the feature values themselves were acquired autonomously. For example, one may define the feature vector to include the 3D pose of each object part involved in an assembly task  \cite{Mollard2015,Niekum2013}. The robot can then use an object localisation method to autonomously obtain the position from sensor data. In this manner, a human may provide the robot with valuable information about the relevant aspects of the environment for the task. However, these human-defined features are generally fixed and may not be optimal for the robot’s purposes. Defining suitable features, as well as mappings from sensory signals to feature values, is also not trivial for more complex tasks, e.g., tasks that involve manipulating deformable objects or varying material properties.

Learning methods provide a useful means for robots to acquire suitable representations autonomously from experience. Deep learning methods in particular have become ubiquitous for their ability to absorb large quantities of data and learn suitable features accordingly. These methods can even be applied to relatively complex scenes with many objects, \emph{e.g.}, pushing around objects in clutter \cite{finn2017deep}. However, even with suitable architectural priors, these approaches often require the robot to obtain vast amounts of experience to learn the features effectively from scratch \cite{pinto2016supersizing}. 

In this paper, we present a method for combining prior task knowledge and experience-based learning to acquire object-centric representations for complex manipulation tasks. To evaluate the proposed approach, we focus on the task of cutting cucumbers and tomatoes into slices. Learning to slice vegetables is a complex task, as it involves manipulating deformable objects into different shapes as well as the creation of new objects in the form of slices. 

\begin{figure*}[tb]
    \centering
    \begin{subfigure}{0.33\linewidth}
    \centering
    \includegraphics[height=0.13\textheight]{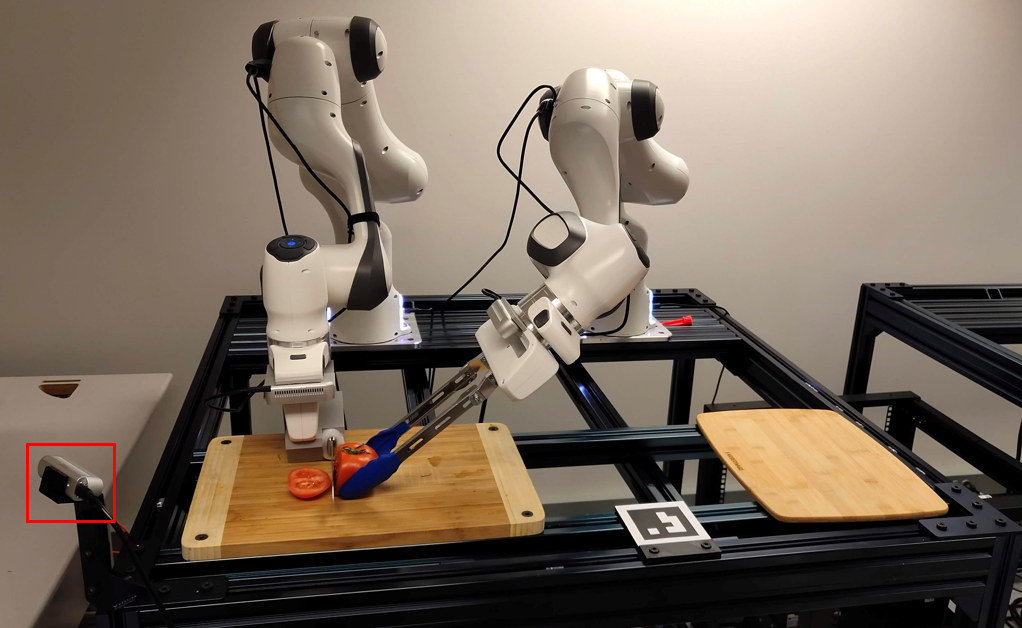}
    \caption{Full view of our setup}
    \label{fig:setup}
    \end{subfigure}%
    \centering
    \begin{subfigure}{0.33\linewidth}
    \centering
    \includegraphics[height=0.13\textheight]{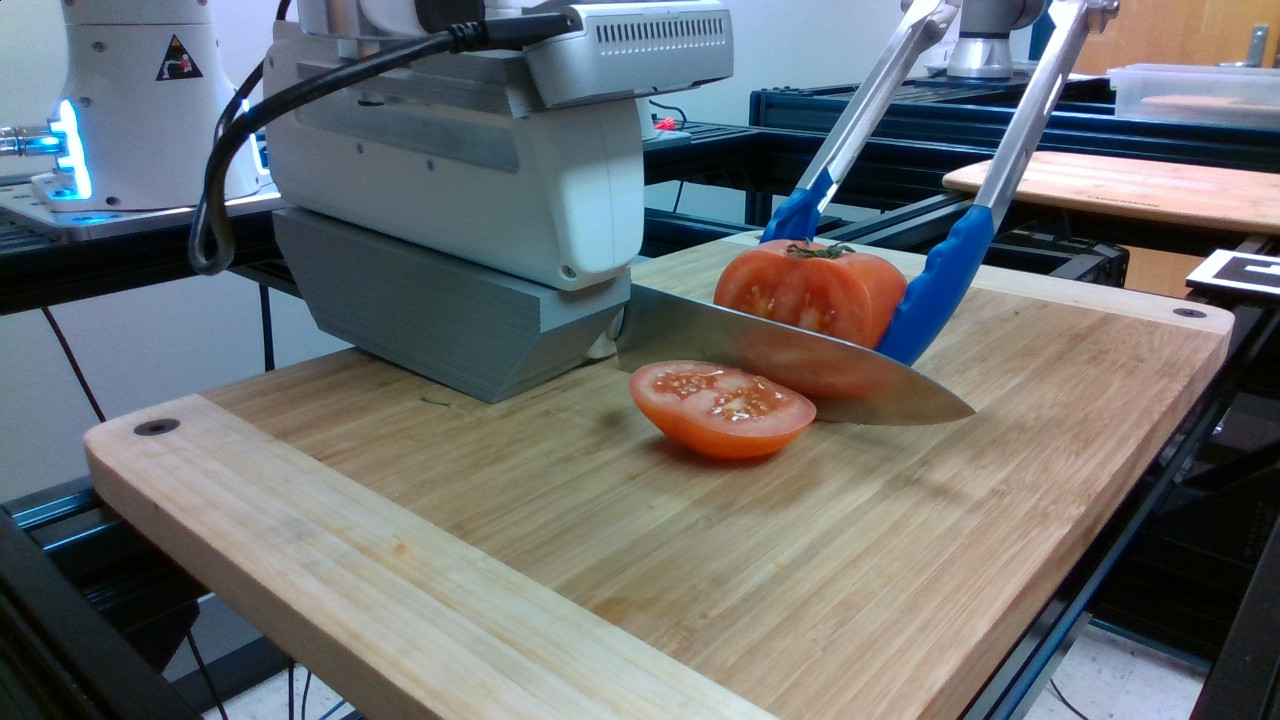}
    \caption{View from side-mounted camera}
    \label{fig:side-camera}
    \end{subfigure}%
    \begin{subfigure}{0.33\linewidth}
    \centering
    \includegraphics[height=0.13\textheight]{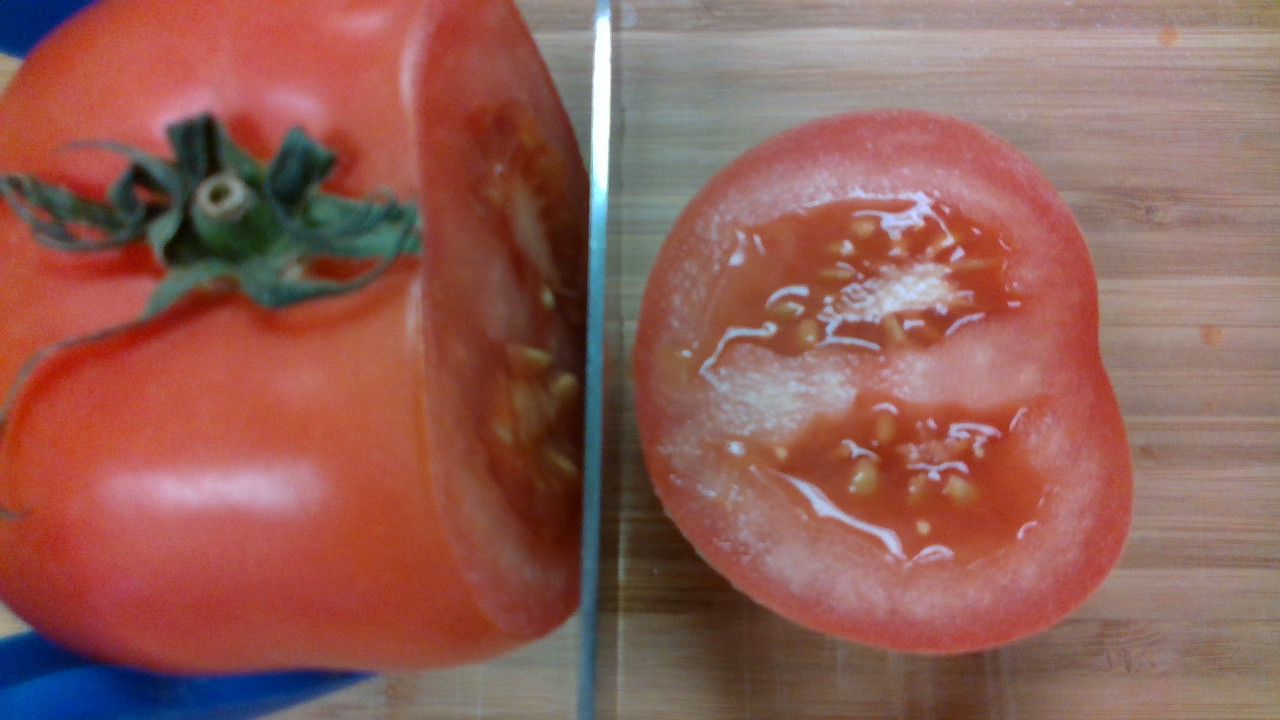}
    \caption{View from arm-mounted camera}
    \label{fig:arm-camera}
    \end{subfigure}%
    
    \caption{Experimental setup with two 7-DOF Franka Emika Panda Research Arms. We have 3 Intel Realsense cameras: one mounted on the side of the robots' structural frame (indicated by the red bounding box in Figure \ref{fig:setup}), one mounted on the cutting arm, and the last mounted on the holding arm. In addition, we have a knife that is grasped by the cutting arm and tongs that are attached to the fingers of the arm holding the vegetable.}
    \vspace{-1em}
\end{figure*}

Figure~\ref{fig:teaser_img} shows an overview of our proposed approach. We use robot interactions to collect training data for the vegetable slicing task.
Introducing meaningful auxiliary tasks while training, allows our model to learn a semantically rich 
embedding space that encodes useful priors and properties, such as thickness of the vegetable being cut, in our state-representation.
This representation is subsequently used to learn a forward model, which allows us to plan in the latent embedding space and use our predictions to further improve our learned representations.

\section{Related Work}

Previous works in robotics have looked into learning visual representations for various tasks \cite{lange2012autonomous, jonschkowski2014state, finn2015learning}. However, most of these methods combine visual representation learning with Reinforcement Learning (RL).
Lange et al. use autoencoders to acquire a state space representation for Q-learning that they then test on low dimensional problems \cite{lange2012autonomous}.
Finn et al. learn state representations from camera images which are then used as input to an RL algorithm with a local time-varying linear model \cite{finn2015learning}.
The representation learning procedures proposed in the above approaches are closely tied with an underlying RL algorithm. 
However, for many real-world robotic tasks, such rewards are either not known, challenging to define, or require domain-specific knowledge.
In contrast, we propose to learn a semantic embedding space through the use of auxiliary tasks which does not assume any known reward function and only utilizes general domain knowledge.


Few works have looked at learning an embedding space for planning using raw pixel inputs \cite{watter2015embed, banijamali2017rce, finn2017deep, ebert2017self, kurutach2018learning}. 
Both, E2C \cite{watter2015embed} and RCE \cite{banijamali2017rce} map input images into a latent embedding space, where they learn locally linear transition models and plan for actions using LQR. However, these methods have only been applied to low dimensional simulated environments such as cartpole and 2-link arm.
Alternately, \cite{finn2017deep, ebert2017self} focus on video prediction models to solve for tasks such as grasping and pushing objects. These methods rely on video input to deduce the spatial arrangement of objects from ambiguous observations, and thus are more focused on representing the visual complexity of the real world.
In contrast, we focus on the problem of learning rich semantic embeddings for object-centered representations. 
Our proposed approach uses an uncalibrated RGB camera and does not require a video input.
We show how using our proposed approach, we are able to plan for multiple steps using a single input image.

Prior works have also looked into the problem of learning an embedding space for predicting the dynamics model for simulated robotics tasks \cite{deisenroth2011pilco, gal2016improving, watter2015embed}. For instance \cite{deisenroth2011pilco} use Gaussian processes to model the forward dynamics,
while \cite{gal2016improving} use neural-network based dynamics models to solve for simple tasks such as two-link balancing.
However, it is not clear how feasible it is to learn the low-level dynamics of slicing objects, which involves multiple deformable objects and complex dynamic interactions.

Finally, few prior works have looked into the problem of learning for cutting and slicing vegetables. In \cite{lenz2015deepmpc} the authors propose DeepMPC for learning to cut a wide range of food items. DeepMPC focuses on learning a model-predictive control algorithm which can be applied for complex non-linear dynamics, such as cutting. 
In contrast, our work focuses on learning suitable, semantically rich representations that aid the overall task of slicing vegetables, and does not focus on the dynamics of the cutting action. 

\begin{figure*}
    \centering
    \includegraphics[width=\textwidth]{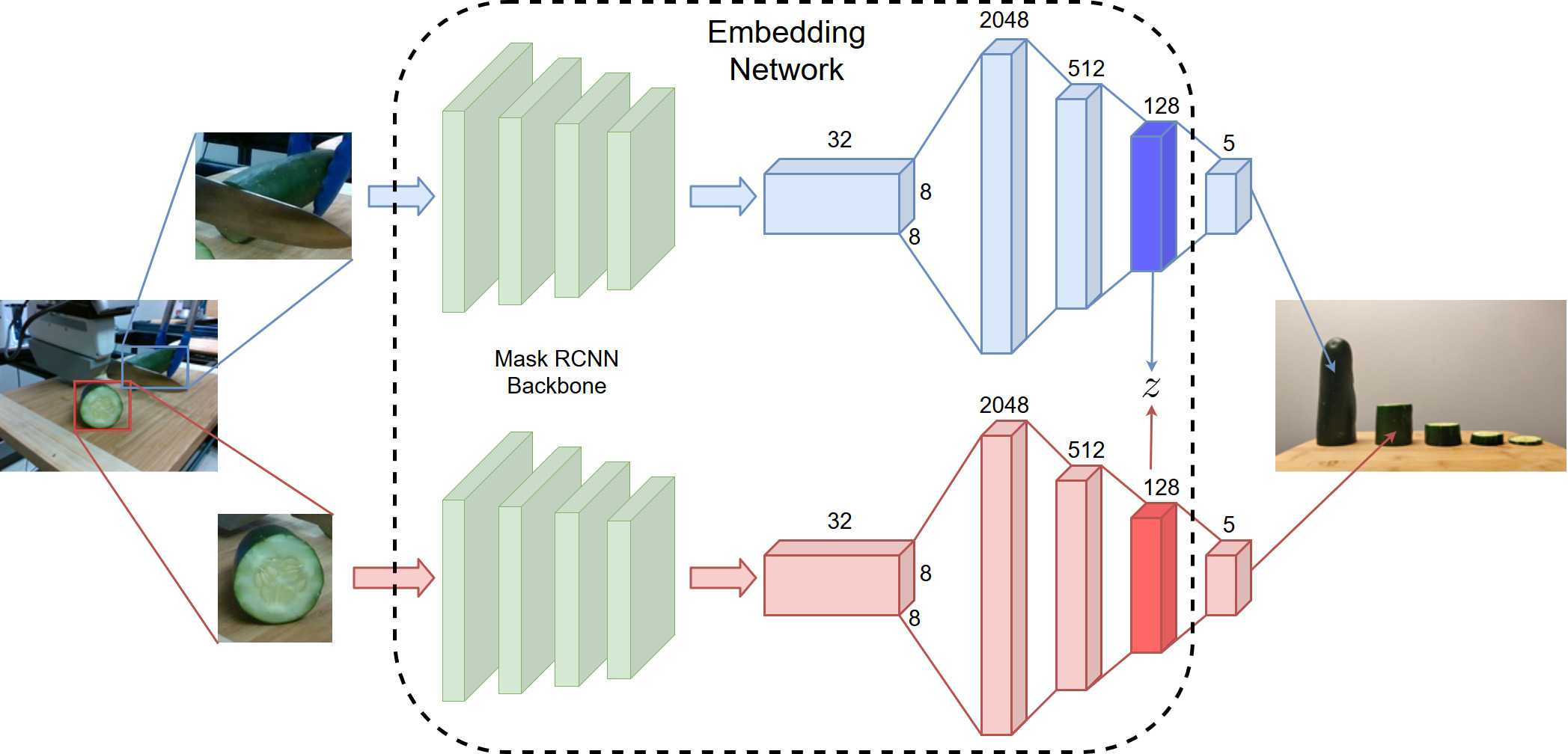}
    \caption{Our Embedding Network is trained using the auxiliary task of predicting the thicknesses of slices from a single raw input image. We use feature maps computed by Mask-RCNN \cite{he2017maskrcnn} backbone network architecture as the input to our embedding network. }
    \label{fig:embedding_arch}
    \vspace{-1em}
\end{figure*}

\section{Robot Slicing Dataset}

In this work, we focus on the problem of a robot slicing vegetables into multiple slices of varying thickness and shape. 
We first discuss the details of our data collection procedure including our experimental setup and the details of learning a motion policy for cutting different vegetables.

\subsection{Experimental Setup}

Figure~\ref{fig:setup} illustrates our experimental setup which consists of a bimanual robotic setup with two 7-DOF Franka Emika Panda Research Arms. The right arm is used to pick, place, and hold the vegetable being cut on the cutting board using tongs attached to its parallel fingers. We refer to this arm as the \textit{holding arm}. The left arm grasps a 3D printed tool holder with an embedded knife that it uses to slice the vegetables held by the other arm. This arm is referred to as the \textit{cutting arm}.

Since we want to learn an embedding space for manipulation using raw pixels, we use a side-mounted Intel RealSense camera (labeled with a red bounding box in Figure~\ref{fig:setup}). We only use the RGB stream from the camera and avoid using any depth information at all. Thus, our setup requires minimal hand-tuning and can be easily replicated. We also have an Intel Realsense camera mounted on each arm, but they were not utilized for this experiment.

\subsection{Parameterized Cutting Action}\label{sec:param_action}

To cut vegetables into slices of different thickness requires the robot to perform multiple varying cutting action. In order to perform this cutting action the robot first needs to detect the end of the vegetable, which it senses using force contact while moving towards the vegetable. The robot then moves up and a certain distance $d$ towards the vegetable to make a slice. It then performs a cutting action which results in a new slice of thickness $\hat{d}$.
To learn the low-level control policy for the cutting action, we use Imitation Learning (IL) in conjunction with Dynamic Motor Primitives (DMPs)~\cite{ijspeert2002movement,schaal2005learning}. We use DMPs in joint space by modeling each joint of the cutting arm using a separate DMP. We collected $10$ trajectories of kinesthetic human demonstrations (humans performing cutting action using the robot arm) to learn the DMP parameters using ridge regression.
We parameterize the above sequence of actions using the thickness $d$ of the cut slice as the main parameter. We refer to this as our parameterized cutting action.

\subsection{Data Collection}
We perform the above learned parameterized cutting action on two different vegetables: cucumbers and tomatoes.
Each has differing spatial properties (cucumbers are longer and thin while tomatoes are shorter and thicker) which leads to contrastive results after completing a slicing task. 

To create our vegetable slicing dataset, we first randomly sample the number of slices to cut at the beginning of each demonstration. For each slice, we randomly sample the slice thickness to cut, ranging from (0.4 cm to 8 cm) for cucumbers and (0.4 cm to 3 cm) for tomatoes. We record the slice thickness for each slice, which we use later while training the forward model. Since we aim to learn directly from raw image inputs we capture an image of the scene (using our side mounted camera) before ($I_t$) and after ($I_{t+1}$) every slice. 
Using this process we are able to collect multiple slices for each given vegetable. Our dataset consists of 50 different demonstrations for cucumbers and 25 demonstrations for tomatoes. Each cucumber demonstration gives us 5 slices on average while a tomato demonstration yields only 3 slices.

To detect relevant objects, such as vegetables, slices, and tools in the scene we fine-tune a pre-trained MaskRCNN model ~\cite{he2017maskrcnn}. We do not use the segmentation part of MaskRCNN instead solely focusing on object detection. To create the fine-tuning dataset we collect $\sim$4000 vegetable images both from Internet (Flickr) and our training setup.

\section{Proposed Approach}

The focus of our work is to learn 
an embedding space for the task of slicing vegetables into multiple pieces of varying thickness. 
To learn such a semantically rich latent embedding space we propose to utilize simple, but related, auxiliary tasks which can be performed directly during robot interactions.
We use the resulting learned representations to learn a forward model for the original task of slicing vegetables. By using the learned forward model to directly plan in the latent embedding space, we demonstrate that our embedding is able to capture the necessary and semantically relevant factors of variations that are important to solve the task of vegetable slicing. 


The above task retains many of the larger challenges posed by the manipulation, such as:
1) Identifying different objects and learning object-centric representations which are amenable for manipulation, 2) Learning a latent representation which captures important properties such as thickness and shape of vegetable slices and can be generalized across different objects or tasks.

\subsection{Learning an Embedding Space}\label{section:embedding_model}

Slicing a vegetable into smaller pieces is a complex task, in that the state of the vegetable being cut changes with every action taken by the manipulator. Additionally, each slice results in a new object being created which further needs to be reasoned about. Moreover, creating multiple new slices of varying thickness requires us to reason 
about many latent factors of the original vegetable such as it's shape and size. This entire process is further conditioned on the type of vegetable being considered, \textit{e.g.} what is considered a thick slice for a tomato considerably differs from a thick slice of a cucumber. 
Thus, we would prefer to learn an embedding space which successfully encodes attributes and properties important for our task such as shape, size and the type of the vegetable being cut. 
Further, our model must accommodate the additional changes in the environment, that includes creation of new slices and update the learned representation accordingly.

In order to learn an embedding space that captures important task attributes 
we pose a classification problem to predict the thickness of the new slice obtained after performing a parameterized cutting action.
In addition to predicting the slice thickness we also predict the thickness of the remaining vegetable.

\subsubsection{Architecture}
Figure~\ref{fig:embedding_arch} shows the proposed network architecture to extract the embedding for an object in the input image. We use the full-scene image, $I$ as input to our finetuned MaskRCNN~\cite{he2017maskrcnn}, such that it predicts a bounding box, $b$ for all the vegetables in the scene.
We then process each object with it's associated bounding box separately.
First, we zoom into each of these bounding boxes (while padding the bounding box with surrounding context). We then pass each of the zoomed-in images through the pre-trained MaskRCNN to collect mid-level image features via the 2D feature maps as given by MaskRCNN's backbone layer \cite{massa2018mrcnn}(see Figure~\ref{fig:embedding_arch}).
We use these feature maps as input to our \emph{embedding network} which consists of a convolution layer followed by multiple fully connected layers. We refer the final output of the fully connected layer as the embedding for a given input. This architecture is referred to as the embedding extractor $\psi$,
that predicts an embedding $z$ for each detected vegetable bounding box.
\begin{equation}
z = \psi(I, b)
\end{equation}
For our particular task of slicing vegetables, we use separate embeddings to represent the main vegetable being cut ($z^o$) and the newly created slice ($z^s$). 

\subsubsection{Auxiliary Tasks}

For our vegetable slicing task, we consider the auxiliary task of predicing the thickness of a new slice and the remaining vegetable, using RGB image as input. 
Predicting the exact thickness value from a raw image is extremely challenging, especially when other objects such as knives, tongs, etc occlude the scene. 
Hence, instead of directly predicting the thickness value, we classify it into $K$ classes, i.e., we predict if the slice is either \emph{very thin, thin, thick, very thick} or \emph{full}. To create these values, we assume a food item with an average thickness (length) $l$ and divide it into the following classes $[0.05l, 0.10l, 0.20l, 0.50l, 1l]$. We use $l=20\text{cm}$ for cucumbers and $l=5cm\text{cm}$ for the tomatoes.
For, the part of the vegetable held by the holding arm we assume $K=4$ classes and do not consider the \textit{very thin} class. This is to account for the fact that a certain minimum thickness is required to hold the food item being cut.

\subsubsection{Loss function}

To train the embedding network we use appropriate loss functions for each auxiliary task. Since we are using thickness classification as our auxiliary task for vegetable slicing we use the cross entropy function as our loss criterion.

\subsection{Learning the Forward Model}\label{section:forward_model}

Once we have learned the initial embedding network, we use the latent space embeddings learn a forward model $\phi$ for planning. 
However, since the embedding network was trained on auxiliary tasks and not the main task, the predicted embeddings may not completely correspond to our forward model task. Fortunately, forward planning in the latent embedding space provides a way to mitigate this issue of task misalignment. Using online forward planning (i.e. planning while performing the task) we can compare the embeddings predicted by the forward model with the true observations which are available to us at no additional cost. This provides us with a useful error signal to improve both our forward model prediction as well as the embedding space learned using auxiliary tasks.
\begin{figure}
    \centering
    \includegraphics[width=0.46\textwidth]{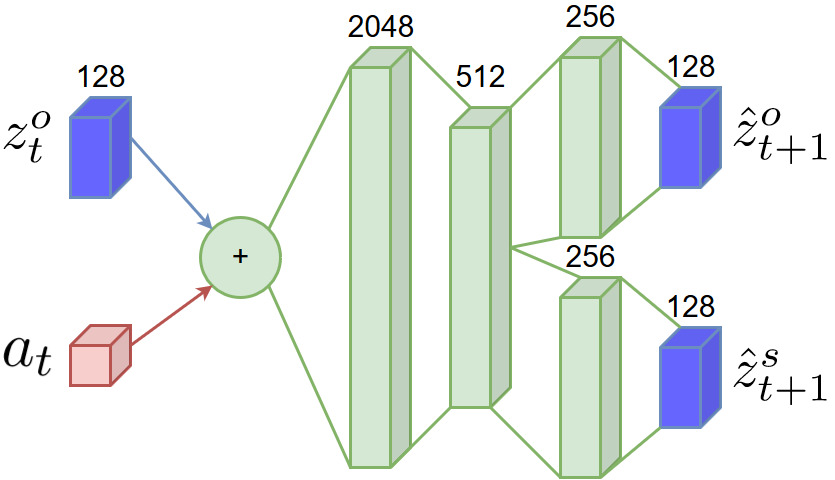}
    \caption{Our Forward Model Architecture predicts the next object and slice embeddings given the current object embedding concatenated with the action input.}
    \label{fig:forward_model_arch}
    \vspace{-1em}
\end{figure}

\subsubsection{Architecture}
Figure~\ref{fig:forward_model_arch} shows the architecture for our forward model. 
Specifically, for the task of slicing vegetables, our forward model predicts the creation of a new slice $\hat{z}^s$ as well as model the change in the original vegetable $\hat{z}^o$.
The input to our forward model consists of the embedding for the original vegetable $z^o_t$ along with the parameterized cutting action $a_t$. 
We train our forward model to predict both the embedding of the remaining vegetable $z^o_{t+1}$ and the embedding of the new created slice $z^s_{t+1}$. To represent if a certain action will not result in the creation of a new cucumber slice we introduce a special STOP symbol embedding which captures the end of the slicing process.

\subsubsection{Multi-Step Forward Model}
The forward model as discussed above only learns using single step predictions. To allow for multi-step prediction in our forward model we use the embedding predicted by our network at time-step $t$ as the input embedding at timestep $t+1$.
In order to ensure that multi-step predictions of the forward model are close to latent embeddings predicted by the embedding network, we propose a curriculum learning approach, where we initially train the forward model for one step predictions only, and increase the planning horizon as the training progresses. We train for the entire sequence towards the end of the training procedure. 

\subsubsection{Loss Functions}

To train the forward model we use the MSE Loss between the prediction error for the embedding i.e. $L_1 = MSE(z^o_{t+1}, \hat{z}_{t+1}^o)$, where $\hat{z}_{t+1}^o = \phi({z}^o_{t}, a_t)$ is the object embedding predicted by the forward model. In addition to the object embedding we also optimize for the slice embedding using $L_2 = MSE(z^s_{t+1}, \hat{z}_{t+1}^s)$, where $\hat{z}_{t+1}^s = \phi({z}^s_t, a_t)$ is the slice embedding predicted by the forward model. 

Thus, the final combined loss for training the forward model can be written as:
\begin{align}
L = \lambda_1\times L_1 + \lambda_2 \times L_2
\end{align}
where $\lambda_{i} ,\ i \in \{1,2\}$ are hyper-parameter values used to weight each loss.

\subsection{Training Details}
For our embedding network architecture (Figure~\ref{fig:embedding_arch}) we use
output embeddings of size $\mathbb{R}^{128}$. For all of our networks we use ReLU as the non-linearity. 
We use the Adam optimizer \cite{kingma2014adam} using a fixed batch size of $16$ during training. To train our forward model for multi-step predictions we use a maximum step length of $5$. We use scheduled sampling to increase the planning horizon with time. For the scheduled sampling process, we initially start with 1-step predictions and increase the planning horizon after every 5 epochs. To finetune MaskRCNN for our vegetable detection task we use the PyTorch implementation of MaskRCNN provided by \cite{massa2018mrcnn}.

\section{Experiments}

In the following sections we demonstrate how utilizing simple auxiliary tasks to train an embedding network results in an informative embedding space. We further show how this learned embedding space affords a rich representation for learning forward models for manipulation. Moreover, we establish that our learned representations can be generalized across difference shapes and sizes, and show the results for our vegetable slicing tasks on two different vegetables, i.e., cucumbers and tomatoes.

\subsection{Embedding Model Results}

\begin{figure}[tb]
    \begin{subfigure}{0.45\columnwidth}
    \centering
    \includegraphics[height=0.19\textheight]{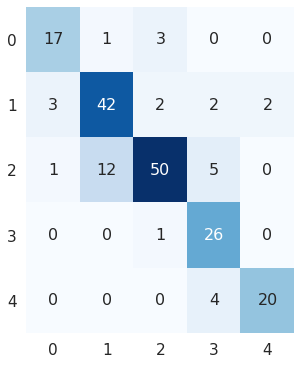}
    \caption{Cucumber slice \\ 
    classification results}
    \end{subfigure}
    \begin{subfigure}{0.45\columnwidth}
    \centering
    \includegraphics[height=0.19\textheight]{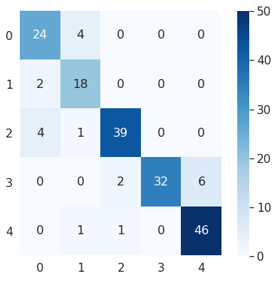}
    \caption{Tomato slice classification results}
    \end{subfigure}
    \caption{Classification results for our auxiliary task of predicting thickness of a slice from a raw image.
     0: \emph{Very Thin}, 1: \emph{Thin}, 2: \emph{Thick}, 3: \emph{Very Thick}, 4: \emph{Full} 
    }
    \label{fig:thickness_classif}
\end{figure}

In order for our approach to learn a robust forward model on the learned embedding, it is imperative that our embedding network performs well on the auxiliary task. 
Therefore, we first look at the classification results for our auxiliary prediction tasks. Figure \ref{fig:thickness_classif} shows the classification results for our embedding network on the task of predicting the thickness of cucumber and tomato slices. The embedding network is able to correctly classify different slices based on their visual appearance for both cucumbers and tomatoes. 

For cucumber slices, we notice that the learned model finds it considerably difficult to discern between \textit{thick} and \textit{thin} classes. This can be attributed to the fact that the actual thickness values of these classes lie substantially close to each other. More importantly, there is little confusion when it comes to predicting the classes that lie further apart from each other, e.g., \textit{very thick} and \textit{very thin} slices. 
Similar results are also observed for tomato slices (see Figure~\ref{fig:thickness_classif} b), i.e., there exists little prediction confusion between thickness classes that are further away from each other.

\begin{figure}[tb]
    \centering
    \begin{subfigure}{0.45\columnwidth}
    \centering
    \includegraphics[height=0.16\textheight]{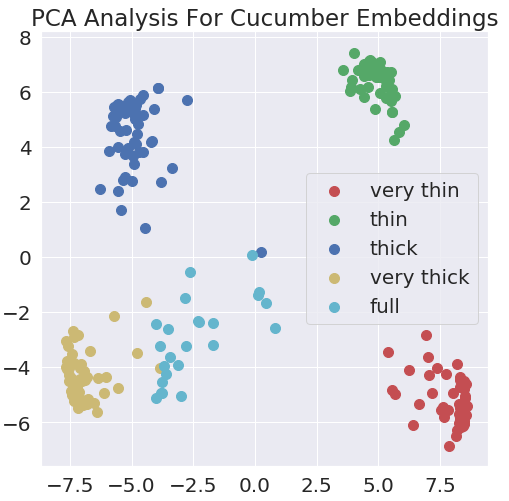}
    \caption{PCA for cucumber \\
    embeddings}
    \end{subfigure}
    \begin{subfigure}{0.45\columnwidth}
    \centering
    \includegraphics[height=0.16\textheight]{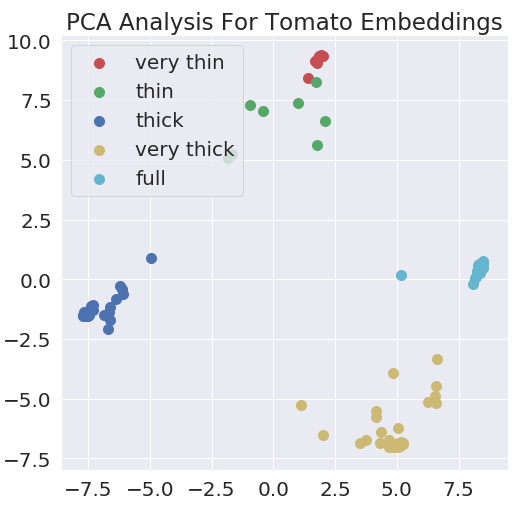}
    \caption{PCA for tomato \\
    embeddings}
    \end{subfigure}
    \caption{PCA analysis for the embeddings learned by our embedding network via thickness classification task on (a) cucumber slices and (b) tomatoes slices. }
    \label{fig:thickness_pca_plot}
    \vspace{-1em}
\end{figure}

\begin{figure*}[tb]
    \centering
    \begin{subfigure}{0.22\linewidth}
    \centering
    \includegraphics[height=0.16\textheight]{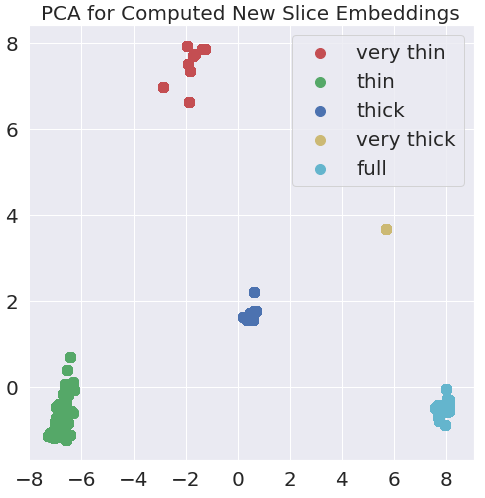}
    \caption{New slice embeddings \\
    computed by embedding \\
    network.}
    \label{fig:computed_new_slice_embedding}
    \end{subfigure}%
    \quad
    \centering
    \begin{subfigure}{0.22\linewidth}
    \centering
    \includegraphics[height=0.16\textheight]{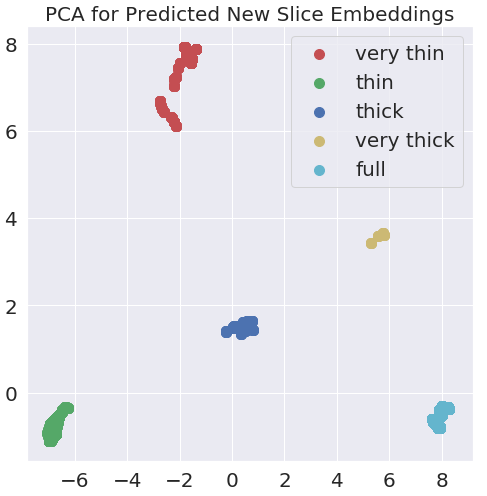}
    \caption{New slice embeddings \\
    predicted by forward model. \\}
    \label{fig:predicted_new_slice_embedding}
    \end{subfigure}%
    \quad
    \begin{subfigure}{0.22\linewidth}
    \centering
    \includegraphics[height=0.16\textheight]{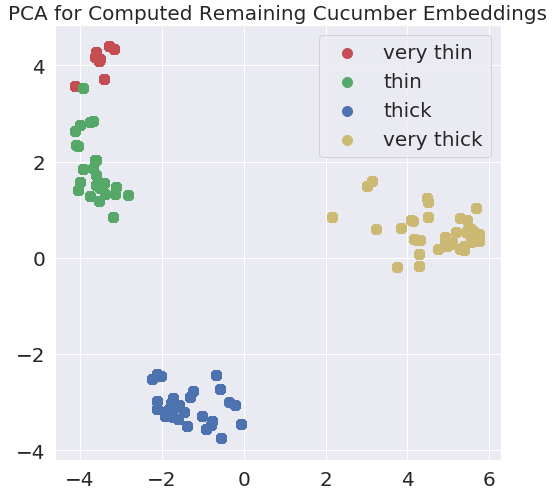}
    \caption{Remaining piece  \\
    embeddings computed by \\
    embedding network.}
    \label{fig:computed_remaining_cucumber_slice_embedding}
    \end{subfigure}%
    \quad
    \begin{subfigure}{0.22\linewidth}
    \centering
    \includegraphics[height=0.16\textheight]{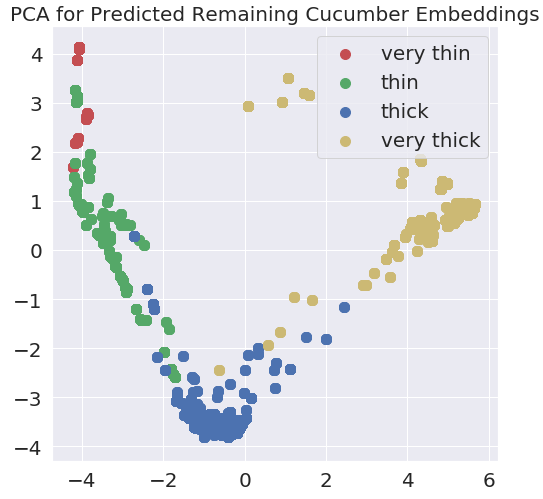}
    \caption{Remaining piece \\
    embedding predicted by \\
    forward model.}
    \label{fig:predicted_remaining_cucumber_slice_embedding}
    \end{subfigure}%
    
    \caption{PCA comparisons for the computed and predicted embeddings of the \textbf{new cucumber slices} and the \textbf{remaining cucumber pieces} after each parameterized cutting action.}
    \label{fig:cucumber_slice_embedding}
\end{figure*}

\begin{figure*}
    \centering
    \begin{subfigure}{0.33\linewidth} 
    \centering
    \includegraphics[height=0.17\textheight]{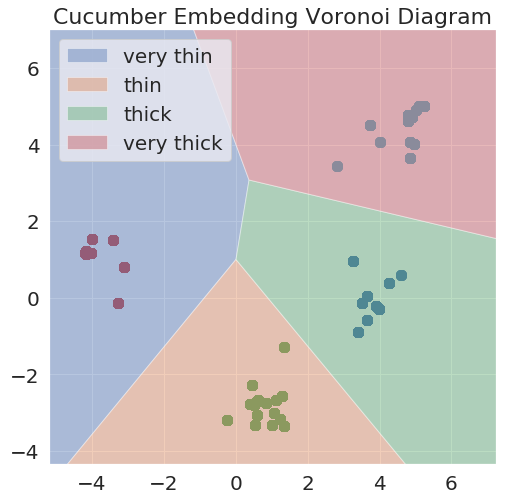}
    \caption{Cucumber Embedding Voronoi Diagram}
    \label{fig:cucumber_voronoi}
    \end{subfigure}%
    \centering
    \begin{subfigure}{0.33\linewidth}
    \centering
    \includegraphics[height=0.17\textheight]{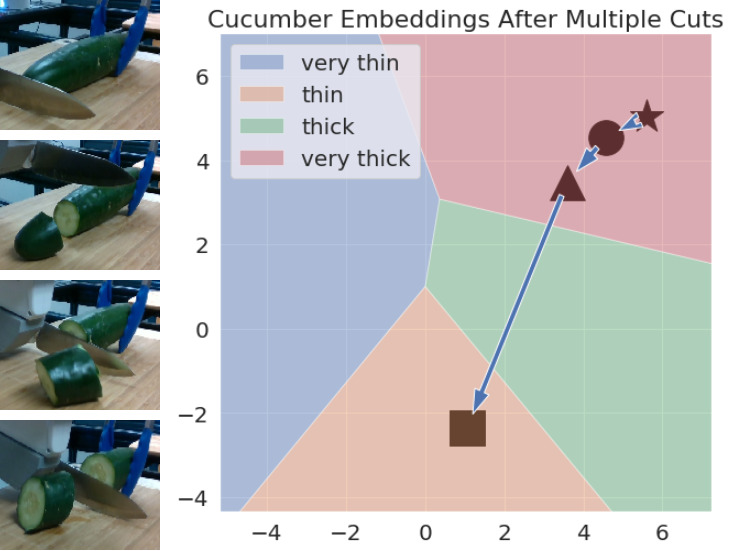}
    \caption{}
    \label{fig:cucumber_embedding_1}
    \end{subfigure}%
    \begin{subfigure}{0.33\linewidth}
    \centering
    \includegraphics[height=0.17\textheight]{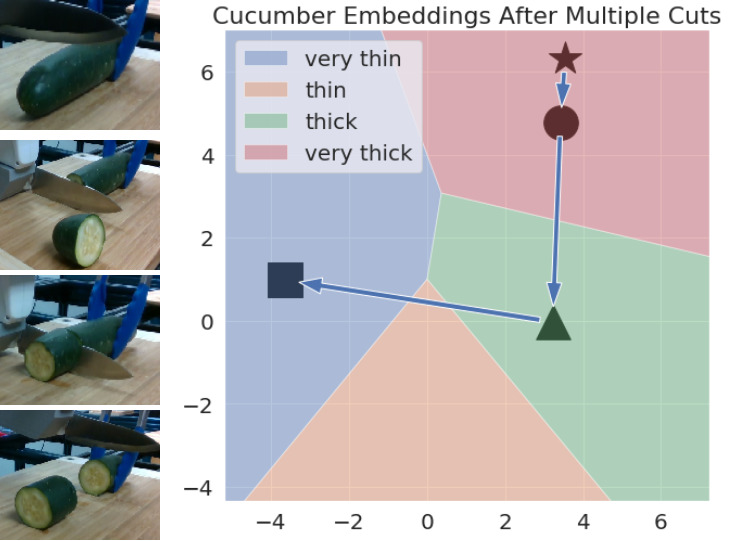}
    \caption{}
    \label{fig:cucumber_embedding_2}
    \end{subfigure}%
    
    \bigskip
    \centering
    \begin{subfigure}{0.33\linewidth}
    \centering
    \includegraphics[height=0.17\textheight]{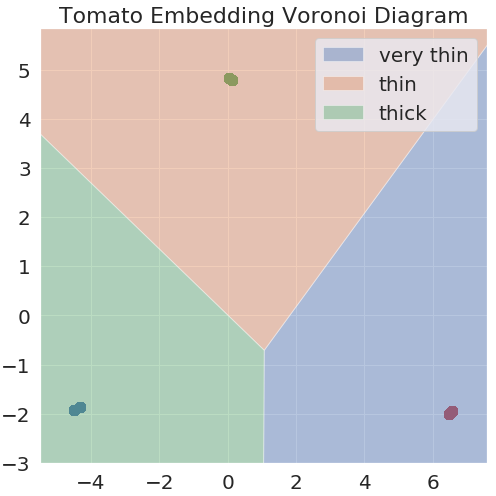}
    \caption{Tomato Embedding Voronoi Diagram}
    \label{fig:tomato_voronoi}
    \end{subfigure}%
    \centering
    \begin{subfigure}{0.33\linewidth}
    \centering
    \includegraphics[height=0.17\textheight]{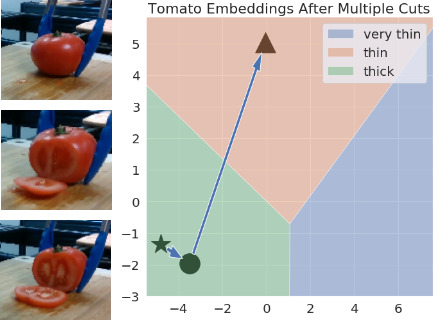}
    \caption{}
    \label{fig:tomato_embedding_1}
    \end{subfigure}%
    \begin{subfigure}{0.33\linewidth}
    \centering
    \includegraphics[height=0.17\textheight]{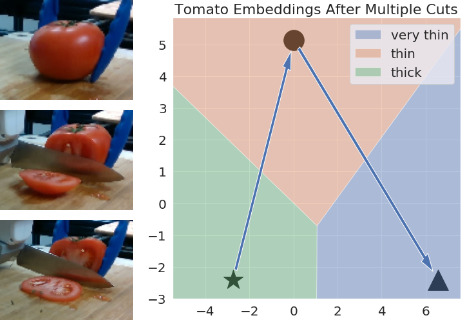}
    \caption{}
    \label{fig:tomato_embedding_2}
    \end{subfigure}%
    
    \caption{
    The above results show how the embeddings predicted by the learned forward model evolve over multiple cutting actions.
    The dots in Figure \ref{fig:cucumber_voronoi} and \ref{fig:tomato_voronoi} represent the \textbf{remaining vegetable piece} embeddings predicted by the embedding network (We only show a few for representation purposes). 
    The $\star$ points in \ref{fig:cucumber_embedding_1}, \ref{fig:cucumber_embedding_2}, \ref{fig:tomato_embedding_1}, and \ref{fig:tomato_embedding_2} denote the initial embeddings while the other shapes represent the predicted embeddings after every step (each cutting action) of the forward model.
    The images to the left of each example visually depict the size of the remaining vegetable piece after each cutting action. 
    }
    
    \vspace{-1em}
\end{figure*}

\begin{figure}
    \centering
    \includegraphics[width=0.46\textwidth]{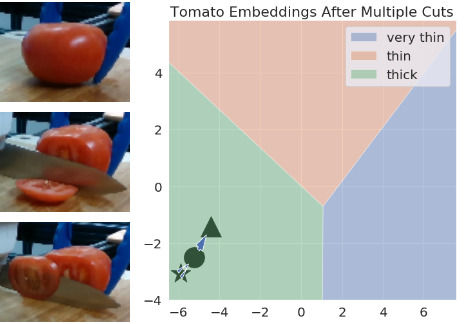}
    \caption{An example of the remaining tomato embeddings changing slowly due to thin slices.}
    \label{fig:tomato_embedding_3}
    \vspace{-1em}
\end{figure}

Given that we have established the efficiency of our trained classifier, we now verify if our model encapsulates and generalizes to the concept of slice thickness. We substantiate this by plotting the embedding space learned by the embedding network, for both the vegetables.
Figure~\ref{fig:thickness_pca_plot} shows the principal component analysis (PCA) on the learned embedding spaces for cucumbers and tomatoes separately. An interesting observation that we would like to impress upon, is that classes which are similar to each other in the visual space, such as \emph{very thick} and \emph{full} (for cucumbers), \emph{very thin} and \emph{thin} (for tomatoes), also lie much closer to each other in each of the embedding spaces. This suggests that our embedding network is able to learn a semantically rich representation which is able to encode properties such as thickness of the object. We now verify whether this learned embedding space allows for useful manipulation by learning a forward model.

\subsection{Learning the Forward Model}
Using our learned embedding network we now look at the embedding results predicted by our forward model. As discussed previously, we use the forward model for multi-step predictions, this forces the forward model to learn from it's own predictions.  
To evaluate the performance of our forward prediction model we first look at the embeddings predicted by our forward model, and compare them to the embeddings computed by our embedding network based on the observed slices and vegetables.

Figure~\ref{fig:cucumber_slice_embedding} visualizes the cucumber embeddings for both our embedding network and the forward model. Additionally, we separately visualize the embeddings for the newly created slice (Figure \ref{fig:computed_new_slice_embedding} and \ref{fig:predicted_new_slice_embedding}) and the embeddings for the remaining cucumber piece (Figure \ref{fig:computed_remaining_cucumber_slice_embedding} and \ref{fig:predicted_remaining_cucumber_slice_embedding}).
The above figure demonstrates that the embeddings learned by the forward model are similar to the embeddings computed using our embedding network. However, it is interesting to note that the embeddings predicted by the forward model show larger variance, when compared to the embeddings computed by the embedding network. We believe this happens because the the forward model learns the semantic meaning associated with the thickness property, which is continuous in nature. This is reinforced by the fact that the above confusion only exists between classes that are close to each other in the visual space, e.g., between \emph{very thin} and thin, \emph{thick} and \emph{very thick}.

\subsection{Multi-Step Planning with Learned Forward Model}
Finally, we also show how combining the above learned embedding space and forward model can be used to plan for the task of vegetable slicing. Figures \ref{fig:cucumber_embedding_1} and \ref{fig:cucumber_embedding_2} show how the embeddings for the remaining cucumber evolve over the course of multiple cutting actions. Similarly, Figures \ref{fig:tomato_embedding_1} and \ref{fig:tomato_embedding_2} shows how the embeddings for the tomato being cut changes over multiple steps. The images on the left of each plot show the observed ground truth images during the interaction. As seen above, for a given input embedding and a parameterized action our forward model is able to correctly simulate how the embedding would be affected after performing the action. For instance, in Figure \ref{fig:tomato_embedding_3} since we cut thin slices the tomato embeddings change slowly within the thick embedding space after 2 steps. In contrast, for Figure \ref{fig:tomato_embedding_2} as we make much thicker slices the embedding quickly changes to very thin, at which point the robot cannot cut a thick slice anymore. For additional results please look at the video at \url{https://sites.google.com/view/learn-embedding-for-slicing/}.

\section{Conclusion}

In this work we proposed a methodology to learn semantically meaningful representations for the task of slicing vegetables. We posit that training an embedding network with simple, albeit related auxiliary tasks, affords representations that capture important task-specific attributes, which aid learning useful forward models for the manipulation task. By introducing the auxiliary task of predicting the thickness of the cut vegetable slice we force our embedding network to model object-centric properties important for the task of slicing vegetables. Moreover, we demonstrate the expressivity of our learned latent embedding space for online planning for the slicing task. We show that in doing so the quality of the representation further improves. Our experiments show that the learned model learns a continuous understanding on important attributes such as thickness of the cut slice.  

\bibliographystyle{IEEEtran} 

\end{document}